\documentclass[10pt,twocolumn,letterpaper]{article}

\usepackage[pagenumbers]{cvpr} 

\usepackage{graphicx}
\usepackage{amsmath}
\usepackage{amssymb}
\usepackage{booktabs}

\usepackage{url}
\usepackage{color}
\usepackage{parskip}
\usepackage{bm}
\usepackage{mathtools}
\usepackage{amsthm}
\usepackage{babel}
\usepackage{multirow}
\usepackage{enumitem}
\usepackage{algorithm2e}
\RestyleAlgo{ruled}

\newcommand{\tabincell}[2]{\begin{tabular}{@{}#1@{}}#2\end{tabular}}

\theoremstyle{remark}
\newtheorem*{rem*}{\protect\remarkname}
\theoremstyle{plain}
\newtheorem{thm}{\protect\theoremname}
\theoremstyle{plain}

\theoremstyle{plain}

\theoremstyle{plain}

\theoremstyle{definition}

\theoremstyle{plain}

\theoremstyle{definition}

\theoremstyle{definition}

\providecommand{\assumptionname}{Assumption}
\providecommand{\corollaryname}{Corollary}
\providecommand{\definitionname}{Definition}
\providecommand{\lemmaname}{Lemma}
\providecommand{\propositionname}{Proposition}
\providecommand{\remarkname}{Remark}
\providecommand{\theoremname}{Theorem}
\providecommand{\conditionname}{Condition}
\providecommand{\examplename}{Example}

\usepackage[pagebackref,breaklinks,colorlinks]{hyperref}

\usepackage[capitalize]{cleveref}
\crefname{section}{Sec.}{Secs.}
\Crefname{section}{Section}{Sections}
\Crefname{table}{Table}{Tables}
\crefname{table}{Tab.}{Tabs.}

\def\cvprPaperID{3348} 
\def\confName{CVPR}
\def\confYear{2022}

\begin{document}

\title{Scalable Penalized Regression for Noise Detection in Learning with Noisy Labels}

\author{Yikai Wang\qquad 
Xinwei Sun\qquad 
Yanwei Fu\thanks{Corresponding author.}\\
School of Data Science, Fudan University\\
{\tt\small 
\{yikaiwang19, sunxinwei, yanweifu\}@fudan.edu.cn
}
}
 
\maketitle

\begin{abstract}
Noisy training set usually leads to the degradation of generalization and robustness of neural networks.
In this paper, we propose using a theoretically guaranteed noisy label detection framework to detect and remove noisy data for Learning with Noisy Labels (LNL).
Specifically, we design a penalized regression to
model the linear relation between network features and one-hot labels, 
where the noisy data are identified by the non-zero mean shift parameters solved in the regression model.
To make the framework scalable to datasets that contain a large number of categories and training data,
we propose a split algorithm to divide the whole training set into small pieces that can be solved by the penalized regression in parallel, 
leading to the Scalable Penalized Regression (SPR) framework.
We provide the non-asymptotic probabilistic condition for SPR to correctly identify the noisy data.
While SPR can be regarded as a sample selection module for standard supervised training pipeline, 
we further combine it with semi-supervised algorithm to further exploit the support of noisy data as unlabeled data.
Experimental results on several benchmark datasets and  real-world noisy datasets show the effectiveness of our framework.
Our code and pretrained models are released at \url{https://github.com/Yikai-Wang/SPR-LNL}.
\end{abstract}

\section{Introduction}
Deep learning has achieved remarkable success on many topics of supervised learning with millions of labeled training data. 
The performance heavily relies on the quality of label annotation since neural networks are susceptible to noisy labels and even can easily memorize randomly labeled annotations~\cite{zhang2017understanding}, leading to the degradation of generalization and robustness. 
In many real-world scenarios, it is expensive and difficult to obtain precise labels, exposing a realistic challenge for supervised deep models to learn with noisy data. 

There is a large literature for this challenge from various perspectives, including modifying the network architectures ~\cite{xiao2015learning,goldberger2017training,chen2015webly,han2018masking} or loss functions~\cite{ghosh2017robust, zhang2018generalized, wang2019symmetric, lyu2020curriculum}, or dynamically selecting clean data during training \cite{song2019selfie, lyu2020curriculum, han2018co, jiang2018mentornet,chen2019understanding,shen2019learning,yu2019does,nguyen2020self}. 
Particularly, the dynamic sample selection methods adopt the spirit of providing only clean data for the training. 
Such a spirit can form a `virtuous' cycle between the noisy data elimination and network training:  
the elimination of noisy data can help the network training; 
and on the other hand, the improved network is empowered with a better ability in picking up clean data. 
As this virtuous cycle evolves, the performance can be improved. 

Typical principles to identify outliers include large loss~\cite{han2018co}, inconsistent prediction~\cite{zhou2021robust}, and irregular feature representation~\cite{wu2020topological}.
The former two principles focus on the label space, while the last one focuses on the feature space of the same class. 
In this paper, we unify the label and feature space 
and assume linear relationship between the feature-label pair (denoted as $(\bm{x}_i,\bm{y}_i)$) of data $i$ by 
\begin{equation}
 \bm{y}_i=\bm{x}_i^{\top}\bm{\beta}+\bm{\varepsilon}, \label{eq1:linear-equation}
\end{equation}
where $\bm{x}_i\in\mathbb{R}^{p}$ is the feature vector, and $\bm{y}_i\in\mathbb{R}^{c}$ is the one-hot label vector; $\bm{\beta}\in\mathbb{R}^{p\times c}$ is the fixed (unknown) coefficient matrix and $\bm{\varepsilon}\in\mathbb{R}^{c}$ is random noise.
This linear relation is approximately established as the networks are trained to minimize the divergence between a (soft-max) linear projection of the feature and  one-hot label vector.
For a well-trained network, the output prediction of clean data is expected to be as similar to a one-hot vector as possible, while for noisy data the output is dense.
Intuitively, when the linear relation is well-approximated without soft-max operation, the corresponding data is likely to be clean data.

The simplest way to identify the suspected outliers in the linear model is checking the predict error, or residual, $\bm{r}_i=\bm{y}_i-\bm{x}_i^{\top}\hat{\bm{\beta}}$, where $\hat{\bm{\beta}}$ is the estimate of $\bm{\beta}$.
The larger $\Vert\bm{r}\Vert$ indicates more possibility for the instance $i$ to be outlier/noisy data.
The classical statistical method to test whether the instance $\bm{r}_i$ is non-zero is using the leave-one-out approach~\cite{weisberg1985applied} to test the externally studentized residual 
\begin{equation}
t_{i}=\frac{\bm{y}_{i}-\bm{x}_{i}^{\top} \hat{\bm{\beta}}_{-i}}{\hat{\sigma}_{-i}\left(1+\bm{x}_{i}^{\top}\left(\bm{X}_{-i}^{\top} \bm{X}_{-i}\right)^{-1} \bm{x}_{i}\right)^{1 / 2}},
\end{equation}
where $\hat{\sigma}$ is the scale estimate and the subscript $-i$ indicates estimates based on the $n-1$ observations, leaving out the $i$-th data where we are testing.
Equivalently, the linear regression model can be re-formulated into explicitly representing the residual by the mean-shift parameter $\bm{\gamma}$ as in~\cite{she2011outlier},
\begin{equation}
\bm{Y}=\bm{X}\bm{\beta}+\bm{\gamma}+\bm{\varepsilon},\quad\varepsilon_{i,j}\sim \mathcal{N}(0, \sigma^2),
\label{eq:lrip}
\end{equation}
where we have the feature $\bm{X}\in\mathbb{R}^{n\times p}$, and label $\bm{Y}\in\mathbb{R}^{n\times c}$ paired and stacked by rows; 
and each row of $\bm{\gamma}\in\mathbb{R}^{n\times c}$, $\bm{\gamma}_i$, represents the predict residual of the corresponding data.
This formulation has been widely studied in different research topics, including economics~\cite{neyman1948consistent, kiefer1956consistency, basu2011elimination, moreira2008maximum}, robust regression~\cite{she2011outlier,fan2018partial}, statistical ranking~\cite{fu2015robust}, face recognition~\cite{wright2009robust}, semi-supervised few-shot learning~\cite{wang2020instance,wang2021trust}, and Bayesian preference learning~\cite{simpson2020scalable}, to name a few.
The focused  formulation is different depending on the specific research tasks. 
For example, for the robust regression problem, the target is to get a robust estimate $\hat{\bm{\beta}}$ against the influence of $\bm{\gamma}$.
Here for solving the problem of learning with noisy labels, we instead aim to \emph{amplify} the impact of $\bm{\gamma}$ such that non-zero values can represent the noisy label that existed in the training set.

To this end, from the statistical perspective, this paper starts from~\cref{eq:lrip} to build up a sample selection framework, dubbed \emph{Scalable Penalized Regression} (SPR), which has theoretical guarantees of consistently identifying noisy data, and thus can efficiently learn with noisy labels.
Naturally, we expect $\bm{\gamma}$ in~\cref{eq:lrip} to be sparse and only a small number of $\bm{\gamma}_i$ are non-zeros, indicating that those data are noisy or outlying.
Thus a sparse penalty is utilized on $\bm{\gamma}_i$ to encourage that the non-zero solution is restricted in a small portion. 
We thus optimize the induced penalized regression problem to solve $\bm{\gamma}$ and identify the instances with non-zero $\bm{\gamma}_i$ as noisy data.
Theoretically, in terms of the model selection consistency theory~\cite{wainwright2009sharp,zhao2006model}, there is some nice statistical property and theoretical insight in our SPR framework, as we can guarantee that, by meeting certain conditions, our SPR should at least in principle, successfully identify all the noisy data.

To incorporate~\cref{eq:lrip} into the  end-to-end training pipeline of deep architecture, the simplest way is to solve~\cref{eq:lrip} for each training mini-batch to detect and remove noisy data. 
However, when we train large model with small batch size, the information of current mini-batch may not be identifiable enough to distinguish true pattern from noise.
On the other hand, use SPR on the whole training data after training an epoch leads to an unacceptable computation cost due to the quadratically increased complexity of solving~\cref{eq:lrip} with the training data.
To design a proper optimization environment for solving~\cref{eq:lrip} that is data-efficient and identifiable, we utilize the whole training set and propose a split algorithm to divide it into small pieces that are class balance with proper data size such that the noisy pattern is \emph{identifiable} and can be solved \emph{efficiently} in parallel, making SPR scalable to large datasets.

Inspired by~\cite{zhou2021learning}, to further encourage the linear relation between features and labels, we propose using a sparse penalty on the fully-connected output before it is soft-maxed.
Moreover, we utilize SPR to train the network in a semi-supervised manner using CutMix~\cite{yun2019cutmix}, regarding the detected noisy data as unlabeled data to fully utilize the feature information.
We conduct extensive experiments to validate the effectiveness of our framework on several benchmark datasets and real-world noisy datasets. 

\textbf{Contributions.} 
Our contributions are as follows:
\begin{itemize}[leftmargin=*,itemsep=0pt,topsep=0pt,parsep=0pt]
\item We present a statistical approach, SPR, to identify noisy data under a general scenario with theoretical guarantees. 
\item A split algorithm is proposed to make SPR scalable to large datasets.
\item A sparse penalty is proposed to encourage the linear relation, and a full training framework that combines SPR with semi-supervised methods is designed.
\item Experiments on benchmark datasets and real-world noisy datasets validate the effectiveness of SPR. 
\end{itemize}

\section{Related Work}
The target of Learning with Noisy Labels (LNL) is to train a more robust model from the noisy dataset.
We can roughly categorize LNL algorithms into two groups: robust algorithm and noise detection.
Robust algorithm does not focus on specific noisy data, but designs specific modules to ensure that networks can be well-trained even from the noisy datasets.
Methods following this direction includes constructing robust network~\cite{xiao2015learning, goldberger2017training,chen2015webly, han2018masking}, robust loss function~\cite{ghosh2017robust, zhang2018generalized, wang2019symmetric, lyu2020curriculum, zhou2021learning, zhou2021asymmetric}, robust regularization~\cite{tanno2019learning,menon2020can, xia2021robust} against noisy labels.

Noise detection method aims to identify the noisy data and design specific strategies to deal with the noisy data, including down-weighting the importance in the loss function for the network training~\cite{thulasidasan2019combating}, re-labeling them to get correct labels~\cite{tanaka2018joint}, or regarding them as unlabeled data in the semi-supervised manner~\cite{Li2020DivideMix}, etc.

For the noise detection algorithm, noisy data are identified by some irregular patterns, including large error~\cite{shen2019learning}, gradient directions~\cite{ren2018learning}, disagreement within multiple networks~\cite{yu2019does}, inconsistency along the training path~\cite{zhou2021robust} and some spatial properties in the training data~\cite{wang2018iterative,lee2019robust,wu2020topological}.
Some algorithms~\cite{veit2017learning, ren2018learning} rely on the existence of an extra clean set to detect noisy data.

After detecting the clean data, the simplest strategy is to train the network using the clean data only or re-weight the data~\cite{patrini2017making} to eliminate the noise.
Some algorithms~\cite{Li2020DivideMix,arazo2019unsupervised} regard the detected noisy data as unlabeled data to fully exploit the distribution support of the training set in the semi-supervised learning manner.
There are also some studies of designing label-correction module~\cite{xiao2015learning,vahdat2017toward,veit2017learning,li2017learning,tanaka2018joint,yi2019probabilistic} to further pseudo-labeling the noisy data to train the network.
Few of those approaches are designed from the statistical perspective with non-asymptotic guarantees.
In this paper, we propose to use SPR to identify the noisy data under general scenarios with statistical guarantees.

\section{Methodology}
\textbf{Problem Formulation.} 
We are given a dataset of image-label pairs $\left\{ \left(\bm{I}_{i},y_{i}\right)\right\} _{i=1}^{n}$, where $\bm{I}_{i}\in\mathcal{I}\subseteq\mathbb{R}^{m},y_{i}\in\mathcal{C}\subseteq\mathbb{R},\left|\mathcal{C}\right|=c$, with the one-hot encoding of $y_{i}$ as $\bm{y}_{i}$. 
We assume that for each instance $i$, $y_{i}$ is corrupted from the ground-truth category $y^\star_{i}$, where the ground-truth and corruption process is unknown.
Our goal is predicting the ground-truth label $y^\star \in \mathcal{C}$ for any $\bm{I} \in \mathcal{I}$, by a neural network composed of a feature extractor $f(\cdot)$ and a classifier $g(\cdot)$.
Typically the network first encodes the image $\bm{I}_{i}$ as a feature vector $\bm{x}_i=f(\bm{I}_{i})$, and return the soft-max probability $\hat{\bm{y}_i}=g(\bm{x}_i)$.

We present our framework -- \emph{Scalable Penalized Regression} (SPR), designed as a sample selection component to the training pipeline of neural networks. 
As mentioned in the introduction, SPR is motivated by the leave-one-out approach~\cite{weisberg1985applied} on the t-test of prediction residuals to identify and remove the noisy data for the network to train, by solving the mean-shift parameter in a sparse linear regression model~(\cref{eq:lrip}). 
Specifically, we use a sparse linear regression model to fit the feature-label pairs $\{\bm{x}_i,\bm{y}_i\}_i^n$ received from the current training time, and solve the corresponding mean-shift parameter $\gamma$ in  
\begin{equation}
\underset{\bm{\beta},\bm{\gamma}}{\mathrm{argmin}}\frac{1}{2}\left\Vert \bm{Y}-\bm{X}\bm{\beta}-\bm{\gamma}\right\Vert _{\mathrm{F}}^{2}+\sum_{i=1}^{n}P\left(\gamma_i;\lambda_i\right),\label{eq:original-problem}
\end{equation}
where $P(\cdot;\cdot)$ is a sparse regularization on $\bm{\gamma}$ with coefficient $\lambda_i$ on row $\bm{\gamma}_i$ to ensure that non-zero $\bm{\gamma}_i$ are sparse, whose corresponding instance are identified as the noisy data.
We denote $\bm{A}_i,\bm{A}_{\cdot,j},\Vert \bm{A} \Vert_{\mathrm{F}}^2:=\sum_{i,j}A^2_{i,j}$ as the $i$-th row, $j$-th column and the square of Frobenius norm, respectively. 

\subsection{Preliminary: Penalized Regression in Statistics}
The penalized regression problem~(\cref{eq:original-problem}) is widely studied in statistics,
where the standard solving algorithm is an alternating optimization pipeline:
for fixed $\bm{\gamma}$, the global optimal solution of $\bm{\beta}$ is the Ordinary Least Square (OLS) estimate of the linear regression problem on $(\bm{X}, \bm{Y}-\bm{\gamma})$;
while for fixed $\bm{\beta}$, the problem is separable in each row of $\bm{\gamma}$, which can be solved by soft-thresholding.
Further, it is shown in~\cite{gannaz2007robust, antoniadis2007wavelet} that the penalized regression problem enjoys the same optimal solution of Huber's M-estimate~\cite{huber2004robust}, which minimizes
\begin{equation}
\underset{\bm{\beta}}{\mathrm{argmin}}
\sum_{i=1}^{n} \rho\left(\frac{\bm{y}_i-\bm{x}_i^{\top}\bm{\beta}}{\sigma};\lambda\right) + \frac{1}{2}cn\sigma,
\label{eq:huber}
\end{equation}
for fixed constants $\sigma>0, c\geq 0, \lambda>0$,
where $\rho(t;\lambda)=t^2/2$ when $|t|\leq\lambda$ and $\rho(t;\lambda)=\lambda |t| -\lambda^2 / 2$ otherwise.
The general formulation of the penalty $P$ can be defined with a three-step construction algorithm introduced in~\cite{she2011outlier}. 
In our experiments, we use the $\ell_1$ norm as the penalty.

\subsection{Penalized Regression for LNL}
In this paper, we regard $\bm{\gamma}$ as the indicator of noisy data, with larger $\Vert \bm{\gamma}_i\Vert$ means more corruption the instance $i$ is suffered. 
We denote $O:=\{i:\Vert \bm{\gamma}_i \Vert \neq 0\}$ as the noisy sample set. 
To estimate $O$, we only need to solve $\bm{\gamma}$ with no need to estimate $\bm{\beta}$.
Thus to simplify the optimization, we substitute the OLS estimate for $\bm{\beta}$  with $\gamma$ fixed into~\cref{eq:original-problem}. 
To ensure that $\hat{\bm{\beta}}$ is identifiable, we apply PCA on $\bm{X}$ to make $p \ll n$ so that the $\bm{X}$ has full-column rank. 
Denote $\tilde{\bm{X}}=\bm{I}-\bm{X}\left(\bm{X}^{\top}\bm{X}\right)^{\dagger}\bm{X}^{\top},\tilde{\bm{Y}}=\tilde{\bm{X}}\bm{Y}$, the~\cref{eq:original-problem} is transformed into
\begin{equation}
\underset{\bm{\gamma}}{\mathrm{argmin}}\frac{1}{2}\left\Vert \tilde{\bm{Y}}-\tilde{\bm{X}}\bm{\gamma}\right\Vert _{\mathrm{F}}^{2}+\sum_{i=1}^{n}P\left(\gamma_i;\lambda_i\right), 
\label{eq:spr}
\end{equation}
which is a standard sparse linear regression for $\bm{\gamma}$. 
Note that in practice we can hardly choose a proper $\lambda$ that works well in all scenarios.
Furthermore, from the equivalence between the penalized regression problem and Huber's M-estimate, the solution of $\bm{\gamma}$ is returned with soft-thresholding.
Thus it is not worth to find the precise solution of a single $\bm{\gamma}$.
Instead, we use a block-wise descent algorithm~\cite{simon2013blockwise} to solve $\bm{\gamma}$ with a list of $\lambda$s and generate the solution path.
As $\lambda$ changes from $\infty$ to $0$, the influence of sparse penalty decreases, and $\bm{\gamma}_i$ are gradually solved with non-zero values, in other words, selected by the model.
Since earlier selected instance is more possible to be noisy, we rank all samples as the descendent order of their selecting time defined as:  
\begin{equation}
C_{i}=\sup\left\{ \lambda:\bm{\gamma}_{i}\left(\lambda\right)\neq0\right\}.
\label{eq:select}
\end{equation}
A large $C_i$ means that the $\bm{\gamma}_i$ is earlier selected. 
Then the top samples are identified as noisy data.

\subsection{Scalable Penalized Regression}
\label{sec:pipeline}
The computation cost of~\cref{eq:spr} is $O(n^2c)$, which increases in quadratic with the growth of the training sample, making it not scalable to large datasets.
Note that we are finding data that are more noisy than others from~\cref{eq:select}; thus we may generate the environment that the noisy instances are easier to be identified and with less computation cost.
To this end, we propose to split the total training dataset into many pieces, each of which contains a small portion of training categories with a small number of training data.
With the splitting strategy, SPR can run on several pieces in parallel and significantly reduce the running time.

Recall that the principle of selecting the group of categories is to reduce the optimizing difficulty and generate an easier environment for finding noisy data.
Our motivation is that categories with less similarity are helpful to identify the noisy data (based on the noisy set recovery theory which we will introduce later),
where the similarity is defined as 
\begin{equation}
\label{eq:simialrity}
s(i,j)=\bm{p}_i^{\top}\bm{p}_j,
\end{equation} 
for class $i,j$ where $\bm{p}$ represents the class prototype.
Specifically, we take the clean features $\bm{x}_i$ of each class extracted by the network along the training iteration, and average them to get the class prototype $\bm{p}_c$ after the current training epoch ends, as
\begin{equation}
\label{eq:prototype}
\bm{p}_c=\frac{\sum_{i=1,y_i=c,i\notin O}^n \bm{x}_i}{\sum_{i=1,y_i=c,i\notin O}^n 1},
\end{equation}
Then the most dissimilar classes are grouped together. In our experiments, we design one group with 10 classes.

For the instances in each group, we split the training data of each class in a balanced way such that each group contains the same number of instances for each class.
The number is determined to ensure that the clean pattern remains the majority in the group, such that optimization can be done easily.
In practice, we select 10 training data of each class to construct the group.
When there is an imbalance between different classes, we use over-sampling strategy to sample the instance of class with less training data multiple times to ensure that each training instance is selected once in some split group.
The detection process is shown in~\cref{alg:spr}.
\begin{algorithm}[ht]
\caption{Scalable Regularized Regression}\label{alg:spr}
\textbf{Input: } Feature matrix $\bm{X}$, label matrix $\bm{Y}$, noisy set $O$.

\textbf{Calculate} prototypes $\bm{P}$ of each class using~\cref{eq:prototype};

\textbf{Divide} classes into most diverged groups based on the similarity within classes using~\cref{eq:simialrity};

\textbf{Split} data of each class in the same group into pieces $\{(\bm{X}_{(i)},\bm{Y}_{(i)})\}$;

\For{\emph{each piece of $(\bm{X}_{(i)},\bm{Y}_{(i)})$ in parallel}}{
\textbf{Solve} $\bm{\gamma}_{(i)}$  using~\cref{eq:spr};

\textbf{Select} the noisy subset $O_{(i)}$ using~\cref{eq:select};
}

\textbf{Group} all the $O_{(i)}$ together and return the result $O$.

\end{algorithm}

\subsection{Learning with Detected Noisy Data}
\textbf{Supervised training manner.}
After estimating the noisy set $O$, the simplest strategy is to remove them and train the network with the remaining clean data.
We show that this strategy will lead to an improvement in testing accuracy.
Note that we assume in~\cref{eq:lrip} that the one-hot encoded label is linearly related to the feature $\bm{X}$; however, in practice, the prediction is obtained via the soft-max function on the $\bm{X} \bm{W}_{\mathrm{fc}}$, where $\bm{W}_{\mathrm{fc}}$ is the weight of the final fully-connected layer (we ignore the bias term for simplicity).

To reduce this gap, inspired by~\cite{zhou2021learning}, we append a $\ell_q$ ($q<1$) penalty on the cross entropy loss, which encourages the linear relationship between $\bm{X}$ and one-hot encoded vector $\bm{Y}$: 
\begin{equation}
\mathcal{L}\left(\bm{x}_i,\bm{y}_i\right) = 1_{i\notin O}(\mathcal{L}_{\mathrm{CE}}\left(\bm{x}_i,\bm{y}_i\right) + \lambda \Vert \bm{x}_i^\top W_{\mathrm{fc}} \Vert_q),
\label{eq:loss-lq}
\end{equation} 
where $q<1$, $\mathcal{L}_{\mathrm{CE}}$ denotes the cross-entropy loss, and $1_{i\notin O}$ denotes the indicator function such that the loss is only calculated on the clean data. 
Note that the $\Vert \bm{x}^\top W_{\mathrm{fc}} \Vert_q$ enforces the $\bm{x}^\top W_{\mathrm{fc}}$ to approximately be one-hot encoded vector as long as $q$ is small enough.
In this training manner, SPR can be regarded as a robust loss function algorithm since we do not modify the training pipeline except the modification of the loss function.

\textbf{Semi-supervised training manner.}
We can further exploit the support of noisy data by incorporating SPR with semi-supervised algorithms.
In this paper, we interpolate part of images between clean data and noisy data as in~\cite{yun2019cutmix},
\begin{subequations}
\label{eq:cutmix}
\begin{align}
\tilde{\bm{x}}&=\bm{M}\odot\bm{x}_{\mathrm{clean}}+(1-\bm{M})\odot\bm{x}_{\mathrm{noisy}}\\
\tilde{\bm{y}}&=\lambda \bm{y}_{\mathrm{clean}}+(1-\lambda)\bm{y}_{\mathrm{noisy}}
\end{align}
\end{subequations}
where $\bm{M}\in\{0,1\}^{W\times H}$ is a binary mask, $\odot$ is element-wise multiplication, and the clean and noisy data are identified by SPR.
Then we train the network using the interpolated data using
\begin{equation}
\mathcal{L}\left(\tilde{\bm{x}},\tilde{\bm{y}}\right) = \mathcal{L}_{\mathrm{CE}}\left(\tilde{\bm{x}},\tilde{\bm{y}}\right).
\label{eq:loss-cutmix}
\end{equation} 
Since $\tilde{\bm{y}}$ is interpolated, it is no longer a one-hot vector, and thus is not sparse.
Hence we do not use the sparse penalty when we train the network using the interpolated data.
Note that SPR is done using the original data without interpolation, hence the linear relationship still holds.
In practice, the above two training method is randomly selected in each mini-batch with the predefined probability.
The full algorithm is shown in~\cref{alg:full}.
\begin{algorithm}[ht]
\caption{Training algorithm}\label{alg:full}
\textbf{Initialize: } Noisy dataset $\{(\bm{I}_i,\bm{y}_i)\}_{i=1}^n$, feature matrix $\bm{X}$, noisy label matrix $\bm{Y}$, noisy set $O=\phi$, CutMix probability $p$.

\For{\emph{ep} from \emph{0} to \emph{total epochs}}
{
\For{\emph{each mini-batch}}
{
Sample $r$ from $U(0,1)$;

\eIf{$r>p$}
{
Train the network using~\cref{eq:loss-lq}.
}
{
Train the network using~\cref{eq:loss-cutmix}.
}
Update $\bm{X}$ visited in the current mini-batch;
}
Run SPR~(\cref{alg:spr}) on $(\bm{X}, \bm{Y})$ and update noisy set $O$;
}
\end{algorithm}

\subsection{Noisy Set Recovery of SPR}
\label{sec:identifiability}
In this part, we provide the result that the~\cref{eq:original-problem} can recover the oracle support set $O$. 
For simplicity, we use the $\ell_1$ norm as the penalty.
In the above we have re-formulate~\cref{eq:original-problem} as~\cref{eq:spr}, which is a standard multi-response regression problem.
Here we further vectorize the problem such that it shares the standard formulation of LASSO.
Then we can use the well-studied model selection consistency result~\cite{zhao2006model,wainwright2009sharp} to support our conclusion.
Specifically, we vectorize $\bm{Y},\bm{\gamma}$ in~\cref{eq:spr} as $\Vec{\bm{y}},\Vec{\bm{\gamma}}$ and the~\cref{eq:spr} turns to
\begin{equation}
\underset{\Vec{\bm{\gamma}}}{\mathrm{argmin}}\frac{1}{2}\left\Vert \Vec{\bm{y}}-\mathring{\bm{X}}\Vec{\bm{\gamma}}\right\Vert _{2}^{2}+\lambda\left\Vert \Vec{\bm{\gamma}}\right\Vert _{1}, 
\label{eq:vec-spr}
\end{equation}
where $\mathring{\bm{X}}=I_{c}\otimes\tilde{\bm{X}}$ with $\otimes$ denoting the Kronecker product operator. 
Denote $S:=\mathrm{supp}(\Vec{\bm{\gamma}}^*)$, then it is sufficient for the recovery of noisy set $O$ to recover $S$. 
We further denote $\mathring{\bm{X}}_{S}(\mathring{\bm{X}}_{S^c})$ as the column vectors of $\mathring{\bm{X}}$ whose indexes are in $S (S^c)$ and $\mu_{\mathring{\bm{X}}}=\max_{i\in S^c}\Vert\mathring{\bm{X}}\Vert_2^2$. 
Then we have
\begin{thm}[Noisy set recovery]
\label{thm:original}
Assume that:\newline 
\emph{C1, Restricted eigenvalue:}
$\quad\lambda_{\min}(\mathring{\bm{X}}_{S}^{\top}\mathring{\bm{X}}_{S})=C_{\min} >0$;\newline 
\emph{C2, Irrepresentability:} there exists a
$\eta \in (0,1]$, such that  $\Vert\mathring{\bm{X}}_{S^c}^{\top}\mathring{\bm{X}}_{S}(\mathring{\bm{X}}_{S}^{\top}\mathring{\bm{X}}_{S})^{-1}\Vert_\infty \leq 1-\eta$;\newline
\emph{C3, Large error:}
$\quad\Vec{\bm{\gamma}}^*_{\min}\coloneqq \min_{i\in S}|\Vec{\bm{\gamma}}^*_{i}|>h(\lambda,\eta,\mathring{\bm{X}},\Vec{\bm{\gamma}}^*)$;\newline
where $\Vert\bm{A}\Vert_\infty\coloneqq \max_i \sum_j |A_{i,j}|$, and 
$h(\lambda,\eta,\mathring{\bm{X}},\Vec{\bm{\gamma}}^*)=\lambda\eta/\sqrt{C_{\min}\mu_{\mathring{\bm{X}}}}+\lambda \Vert(\mathring{\bm{X}}_{S}^{\top}\mathring{\bm{X}}_{S})^{-1}\mathrm{sign}(\Vec{\bm{\gamma}}_{S}^*)\Vert_\infty$.\newline 
Let $\lambda\geq \frac{2\sigma \sqrt{\mu_{\mathring{\bm{X}}}}}{\eta}\sqrt{\log cn}$. 
Then with probability greater than $1-2(cn)^{-1}$, model~\cref{eq:vec-spr} has a unique solution $\hat{\Vec{\bm{\gamma}}}$ such that: 1) If C1 and C2 hold, $\hat{O}\subseteq O$;2) If C1, C2 and C3 hold, $\hat{O}= O$.
\end{thm}
Note that The Theorem 1 is extended from the  model selection consistency in~\cite{wainwright2009sharp}, which only provides the conclusion that $\hat{S}\subseteq S$ and $\hat{S}= S$, respectively.
Here we show that $\hat{S}\subseteq S$ leads to $\hat{O}\subseteq O$, and of course $\hat{S}= S$ leads to $\hat{O}= O$ in our case.
For instance $i$, $i\in O^c$ only when $\gamma_{i,j}= 0$ for all $j$, then all the vectorized indexes are in $S^c$.
When $\hat{S}\subseteq S$, all vectorized indexes of instance $i$ are in $\hat{S}^c$, which means $i\in\hat{O}^c$ and leads to $\hat{O}\subseteq O$.

C1 is necessary to get a unique solution, and in our case is mostly satisfied with the nature assumption that  the clean data is the majority in the training data.
If C2 holds, the estimated noisy data is the subset of truly noisy data.
This condition is the key to ensuring the success of SPR, which requires divergence between clean and noisy data such that we cannot represent clean data by noisy data.
If C3 further holds, the estimated noisy data is exactly all the truly noisy data.
C3 requires the error measured by $\gamma_i$ is large enough to be identified from random noise.
If the conditions fail, SPR will fail in a non-vanishing probability, not deterministic.

\begin{table*}[ht] 
\centering
\begin{tabular}{l|l|cccc|ccc}
\toprule 
\multirow{2}{*}{Dataset} & \multirow{2}{*}{Method} &  \multicolumn{4}{c|}{Sym. Noise Rate} &  \multicolumn{3}{c}{Asy. Noise Rate}\tabularnewline
& & 0.2 & 0.4 & 0.6 & 0.8 & 0.2 & 0.3 & 0.4 \tabularnewline
\midrule
\midrule
\multirow{8}{*}{\tabincell{l}{MNIST\\(C2F2)}}
& CE & 91.6 & 74.0 & 49.4 & 22.7 & 94.6 & 88.8 & 82.3 \tabularnewline
& FL & 91.7 & 74.5 & 50.4 & 22.7 & 94.3 & 89.1 &82.1 \tabularnewline
& GCE & 98.9 & 97.2 & 81.5 & 34.0 & 96.7 & 89.1 & 81.5 \tabularnewline
& SCE & 98.9 & 97.4 & 88.8 & 48.8 & 98.0 & 93.7 & 85.4 \tabularnewline
& NLNL & 98.3 & 97.8 & 96.2 & 86.3 & 98.4 & 97.5 & 95.8 \tabularnewline
& APL & 99.1 & 98.4 & 95.7 & 73.0 & 98.9 & 96.9 & 91.5 \tabularnewline
& SR & 99.2 & 99.2 & 98.9 & 98.0 & 99.3 & 99.2 & 99.2 \tabularnewline
\cline{2-9}
& SPR & \textbf{99.3} & \textbf{99.2} & \textbf{99.2} & \textbf{98.7} & \textbf{99.3} & \textbf{99.2} & \textbf{99.2} \tabularnewline
\midrule
\midrule
\multirow{15}{*}{\tabincell{l}{CIFAR-10\\(ResNet-18)}}
& Standard & 85.7 & 81.8 & 73.7 & 42.0 & 88.0 & 86.4 & 84.9 \tabularnewline
& Forgetting & 86.0 & 82.1 & 75.5 & 41.3 & 89.5 & 88.2 & 85.0 \tabularnewline
& Bootstrap & 86.4 & 82.5 & 75.2 & 42.1 & 88.8 & 87.5 & 85.1 \tabularnewline
& Forward & 85.7 & 81.0 & 73.3 & 31.6 & 88.5 & 87.3 & 85.3 \tabularnewline
& Decoupling & 87.4 & 83.3 & 73.8 & 36.0 & 89.3 & 88.1 & 85.1 \tabularnewline
& MentorNet & 88.1 & 81.4 & 70.4 & 31.3 & 86.3 & 84.8 & 78.7 \tabularnewline
& Co-teaching & 89.2 & 86.4 & 79.0 & 22.9 & 90.0 & 88.2 & 78.4 \tabularnewline
& Co-teaching+ & 89.8 & 86.1 & 74.0 & 17.9 & 89.4 & 87.1 & 71.3 \tabularnewline
& IterNLD & 87.9 & 83.7 & 74.1 & 38.0 & 89.3 & 88.8 & 85.0 \tabularnewline
& RoG & 89.2 & 83.5 & 77.9 & 29.1 & 89.6 & 88.4 & 86.2 \tabularnewline
& PENCIL & 88.2 & 86.6 & 74.3 & 45.3 & 90.2 & 88.3 & 84.5 \tabularnewline
& GCE & 88.7 & 84.7 & 76.1 & 41.7 & 88.1 & 86.0 & 81.4 \tabularnewline 
& SCE & 89.2 & 85.3 & 78.0 & 44.4 & 88.7 & 86.3 & 81.4 \tabularnewline
& TopoFilter & 90.2 & 87.2 & 80.5 & 45.7 & 90.5 & 89.7 & 87.9 \tabularnewline
\cline{2-9}
& SPR & \textbf{93.2} & \textbf{91.0} & \textbf{82.7} & \textbf{64.1} & \textbf{92.8} & \textbf{91.3} & \textbf{89.0}\tabularnewline
\bottomrule
\end{tabular}
\vspace{-0.1in}
\caption{Test accuracies on several benchmark datasets with different settings.
The best result is \textbf{boldfaced}.
Results of competitors on MNIST are reported in~\cite{zhou2021learning}, and on CIFAR10 are reported in~\cite{wu2020topological}.
\label{tab:main-results} }
\vspace{-0.1in}
\end{table*}

\section{Experiments}
\textbf{Datasets.}
We validate the effectiveness of SPR on synthetic noisy datasets MNIST~\cite{lecun1998gradient} and CIFAR10~\cite{krizhevsky2009learning}, and real-world noisy datasets ANIMAL10~\cite{song2019selfie} and WebVision~\cite{li2017webvision}.
We consider two types of noisy labels for MNIST and CIFAR10:
(i) Symmetric noise:
Every class is corrupted uniformly with all other labels;
(ii) Asymmetric noise:
Labels are corrupted by similar (in pattern) classes. The ANIMAL10 is published with mislabeling (the ratio is 8\%) and the corruption process and noise type in ANIMAL10 are unknown. 
WebVision has 2.4 million images collected from the internet with the same category list with ImageNet ILSVRC12.
Thus, the ANIMAL10 and WebVision datasets can be regarded as a real-world challenge. 

\textbf{Backbones}. 
For MNIST, we use two convolutional layers followed by two fully-connected layers, denoted as \textit{C2F2}. For CIFAR10, a ResNet-18~\cite{he2016deep} network is utilized. 
For ANIMAL10 we use VGG19-BN~\cite{simonyan2015very} as our backbone.
And for WebVision we use Inception-ResNet~\cite{szegedy2017inception} to extract features.

\textbf{Hyperparameter setting}.
We use SGD to train all the networks with momentum 0.9 and a cosine learning rate decay strategy.
The initial learning rate is set as 0.1 for ANIMAL10 and 0.01 for others.
The weight decay is set as 1e-3,1e-4, 1e-3, 5e-4 for MNIST, CIFAR10, ANIMAL10, and WebVision, respectively.
We use a batch size of 128 for all experiments.
We use random crop and random horizontal flip as augmentation strategies for CIFAR10, ANIMAL10, and WebVision.
The network is trained for 50 epochs for MNIST, 180 epochs for CIFAR10, 160 epochs for ANIMAL10, and 300 epochs for WebVision.
We use $q=0.2$ in~\cref{eq:loss-lq} with coefficient $\lambda$ of the sparse penalty initialized as 1.2 and is increased by multiplying 1.2 for MNIST, and 1.02 for others.
In CIFAR10 with noise rate 0.8, we do not increase the $\lambda$.
We simply select half of the training data as noisy data in all of our experiments.

\subsection{Evaluation on Synthetic Label Noise}
\textbf{Competitors}. 
In this part, we first use SPR with only using the supervised training manner on MNIST to compare with robust loss function methods. 
Then we use the full SPR model on CIFAR-10 to compare with sample selection algorithms and other popular algorithms.
We use cross-entropy loss (CE) as baseline algorithm for two datasets.
For MNIST, we also compare with competitors including
an effective loss function Focal Loss (FL)~\cite{lin2017focal},
some refined algorithms for CE loss like GCE~\cite{zhang2018generalized} and SCE~\cite{wang2019symmetric}, NLNL~\cite{kim2019nlnl} which utilizes complementary labels against the noise,
APL~\cite{ma2020normalized} which combines robust active and passive loss to train the network.
SR~\cite{zhou2021learning} which utilizes the sparse regularization combined with the feature normalization and temperature scaling method to train the network. 
For CIFAR-10, we compare SPR with algorithms include 
Forgetting~\cite{arpit2017closer} with train the network using dropout strategy, 
Bootstrap~\cite{reed2015training} which train with bootstrapping, 
Forward Correction~\cite{patrini2017making} which corrects the loss function to get a robust model, 
Decoupling~\cite{malach2017decoupling} which uses a meta update strategy to decouple the update time and update method, 
MentorNet~\cite{jiang2018mentornet} which uses a teacher network to help train the network, 
Co-teaching~\cite{han2018co} which uses two networks to teach each other, 
Co-teaching+~\cite{yu2019does} which further uses an update by disagreement strategy to improve Co-teaching, 
IterNLD~\cite{wang2018iterative} which uses an iterative update strategy, 
RoG~\cite{lee2019robust} which uses generated classifiers, 
PENCIL~\cite{yi2019probabilistic} which uses a probabilistic noise correction strategy, 
GCE~\cite{zhang2018generalized} and SCE~\cite{wang2019symmetric} which are extensions of standard cross-entropy loss function, 
and TopoFilter~\cite{wu2020topological} which uses feature representation to detect noisy data.
For each dataset, all the experiments are running with the same backbone to make a fair comparison.

As in~\cref{tab:main-results},
SPR enjoys a high performance compared with other robust loss function algorithm without using noisy data on MNIST, and shows a high superiority to many competitors on CIFAR-10, validating the effectiveness of SPR on different noise scenarios.

\subsection{Evaluation on Real-World Noisy Datasets} 
In this part, we compare SPR with other methods in  real-world noisy datasets including ANIMAL10 and WebVision. 
For WebVision we use CutMix probability of $1.0$.

\textbf{Competitors}. 
For ANIMAL10, we compare with the baseline of directly training with cross-entropy loss (CE), as well as previous works including Nested(ND), CE + Dropout (CED), SELFIE~\cite{song2019selfie}, PLC~\cite{zhang2021learning}, and NestedCoTeaching (NCT)~\cite{chen2021boosting}.
For WebVision, we compare with directly training with cross-entropy loss (CE), as well as Decoupling~\cite{malach2017decoupling}, D2L~\cite{ma2018dimensionality}, MentorNet~\cite{jiang2018mentornet}, Co-teaching~\cite{han2018co}, Iterative-CV~\cite{chen2019understanding}, and DivideMix~\cite{Li2020DivideMix}.

The results of real-world datasets are shown in~\cref{tab:real-world}, where the results of CE and SELFIE on ANIMAL is reported in~\cite{song2019selfie},
the results of ND, CED, and NCT is reported in~\cite{chen2021boosting},
while the result of PLC is reported in their paper.
The results of competitors on WebVision are reported in~\cite{Li2020DivideMix} and the result of CE is reported in~\cite{zhou2021learning}.
Our algorithm SPR enjoys superior performance to all the competitors, showing the ability of handling real-world challenges.

\begin{table}[ht]
\centering
\begin{tabular}{lc|lc}
\toprule
\multicolumn{2}{c|}{ANIMAL10} & \multicolumn{2}{c}{WebVision} \tabularnewline 
Model & Accuracy & Model & Accuracy \tabularnewline
\midrule
\midrule
CE & 79.4 & CE  & 66.96\tabularnewline
Nested & 81.3 & Decoupling & 62.54\tabularnewline
CED & 81.3 & MentorNet & 63.00\tabularnewline
SELFIE & 81.8 & Co-teaching & 63.58 \tabularnewline
PLC & 83.4 & Iterative-CV & 65.24\tabularnewline
NCT & 84.1 & DivideMix & 77.32\tabularnewline
\midrule
SPR & \textbf{86.8} & SPR & \textbf{78.12}\tabularnewline
\bottomrule
\end{tabular}
\vspace{-0.1in}
\caption{Results on real-world datasets. 
The best are in \textbf{bold}. \label{tab:real-world}}

\vspace{-0.15in}
\end{table}

\subsection{More Analysis of SPR }
\label{sec:analysis}
\begin{figure*}[!ht]
\centering 
\begin{subfigure}{0.3\linewidth}
\centering
\includegraphics[width=1\linewidth]{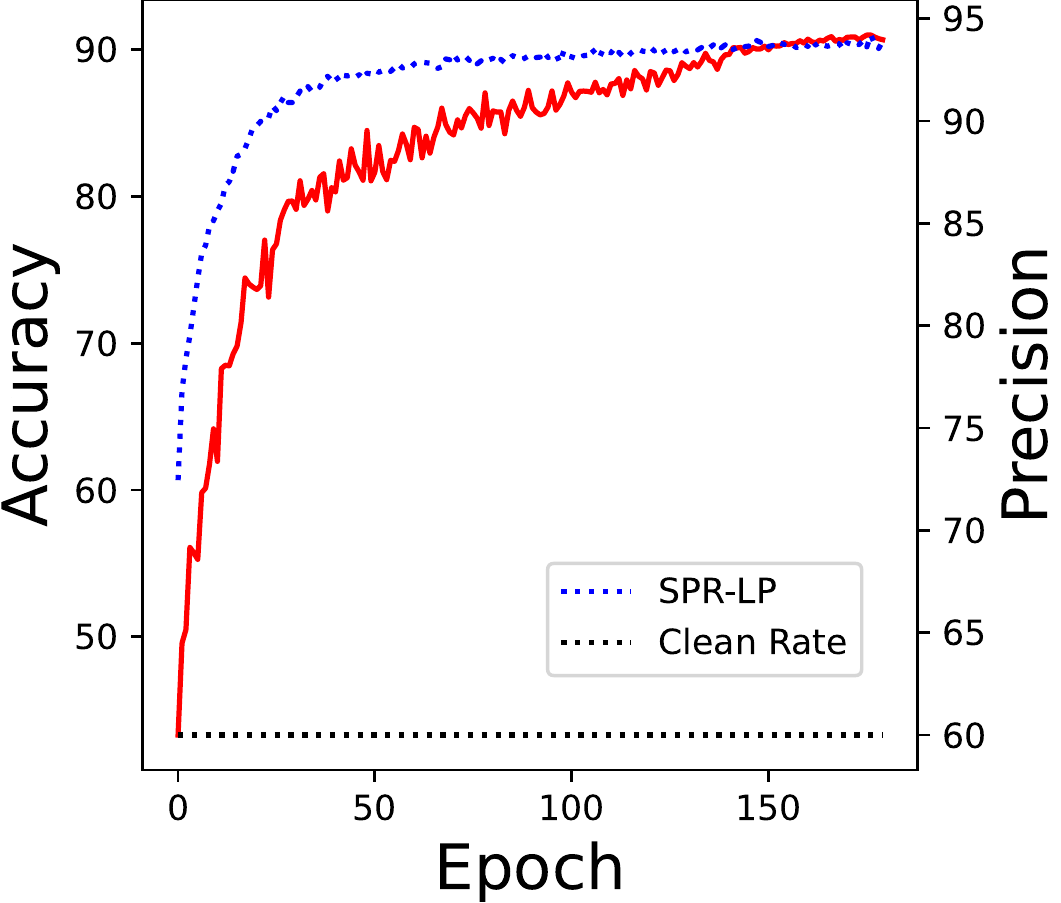} 
\caption{Symmetric-40\%}
\end{subfigure}
\begin{subfigure}{0.3\linewidth}
\centering
\includegraphics[width=1\linewidth]{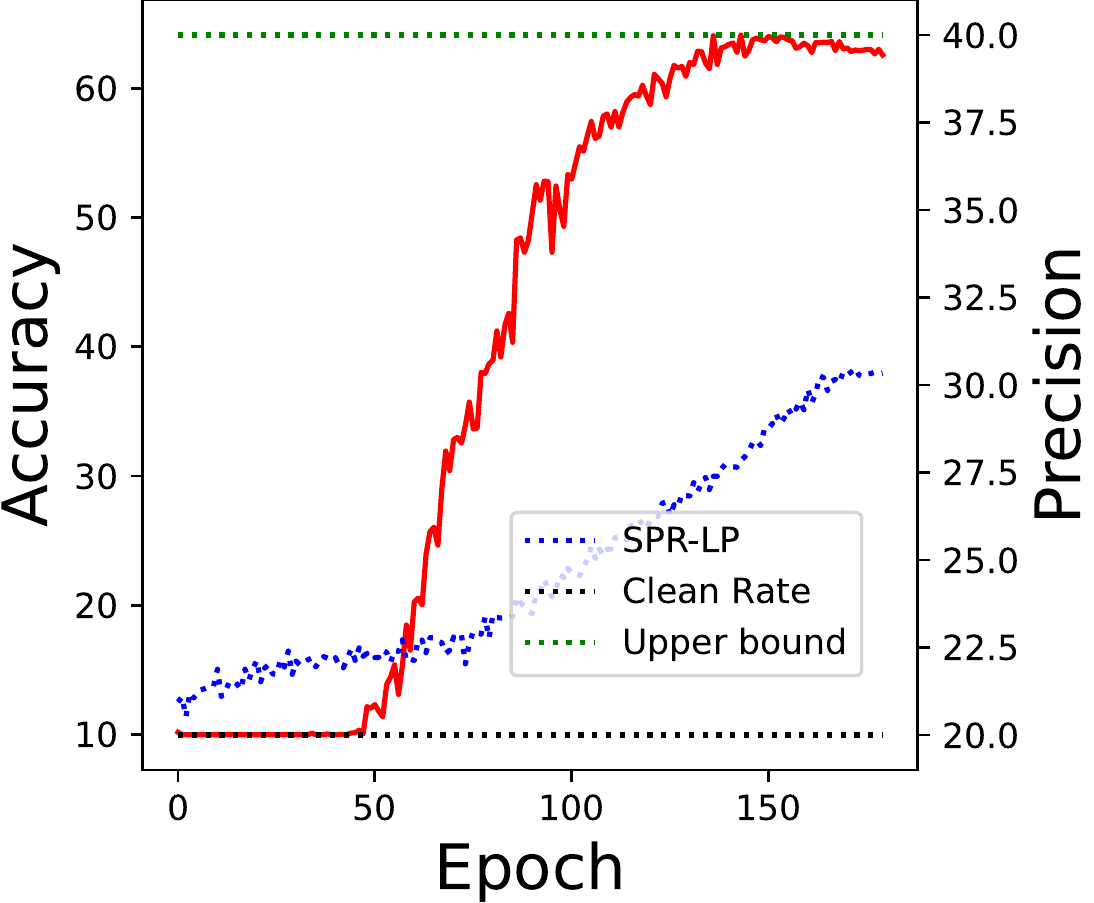} 
\caption{Symmetric-80\%}
\end{subfigure}
\begin{subfigure}{0.3\linewidth}
\centering
\includegraphics[width=1\linewidth]{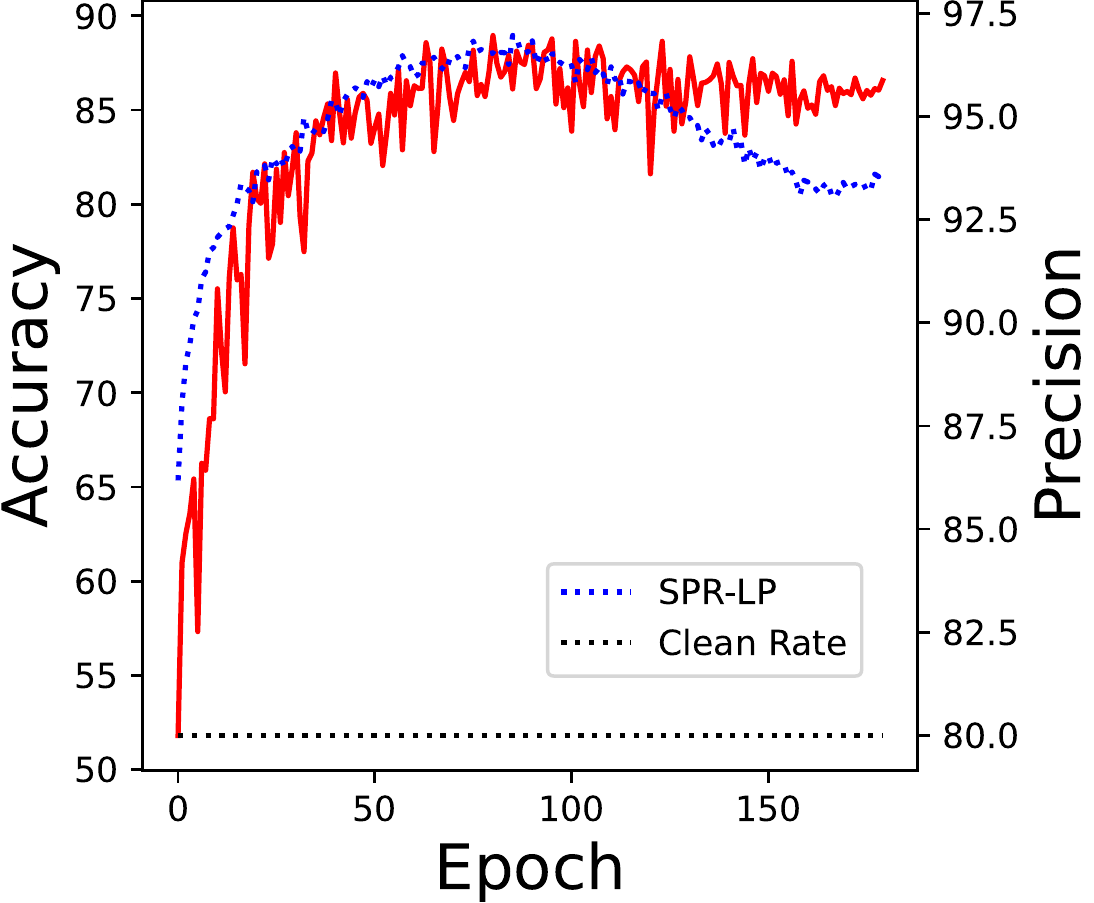} 
\caption{Asymmetric-40\%}
\end{subfigure}
\vspace{-0.1in}
\caption{Accuracy and Label precision of SPR under different noise scenarios.
The red line is the accuracy of SPR, while the dotted line is the label precision. \label{fig:precision}
}

\vspace{-0.15in}
\end{figure*}

\textbf{Precision of noisy detection}.
Besides accuracy, another metric to test the capacity of a sample selection algorithm is \textit{label precision}: 
the ratio of true clean labels in the detected clean instances.
In this part, we check the label precision of SPR to show the sample selection effectiveness.
We conduct our experiments on the symmetric noise rate of 0.4 and 0.8, as well as asymmetric noise rate of 0.4.
Results are shown in~\cref{fig:precision}.
SPR enjoys a monotonically increasing label precision in the symmetric noise setting, leading to a better training environment than the standard noisy dataset.
When the training process ends, almost all the selected training data is guaranteed to be clean data (93.90\% in the symmetric-40\% setting).
In the symmetric-80\% setting, due to the strategy of selecting half of the training data, the upper bound of the precision is 40\%, as illustrated.
In this high noise rate scenario, SPR can still achieve the precision of 30.34\%, which means that 76.24\% of the clean training instances are detected by our algorithm.
Note that in the asymmetric-40\%, the label precision is first increased then decreased, and ends with 93.50\%.
Though it is still high, the accuracy drops with the label precision, suggesting that an early stopping strategy is needed in the asymmetric noisy setting.
We leave it as a future work to provide a fine-grained framework for different noise scenarios.

\begin{table}
\centering
\begin{tabular}{lc}
\toprule
Model & Accuracy\tabularnewline
\midrule
\midrule
CE & 65.5\tabularnewline
CE + SPR & 80.4\tabularnewline
CE + $\ell_q$ & 71.6\tabularnewline
CE + CutMix & 87.0\tabularnewline
CE + SPR + $\ell_q$ & 88.5\tabularnewline
CE + SPR + CutMix & 89.2\tabularnewline
\midrule
Full & 91.0\tabularnewline
\bottomrule
\end{tabular}
\vspace{-0.1in}
\caption{Accuracy of using different modules in SPR.\label{tab:ablation}}
\vspace{-0.1in}
\end{table}

\begin{figure}[!ht]
\centering 
\begin{subfigure}{0.45\columnwidth}
\centering
\includegraphics[width=1\linewidth]{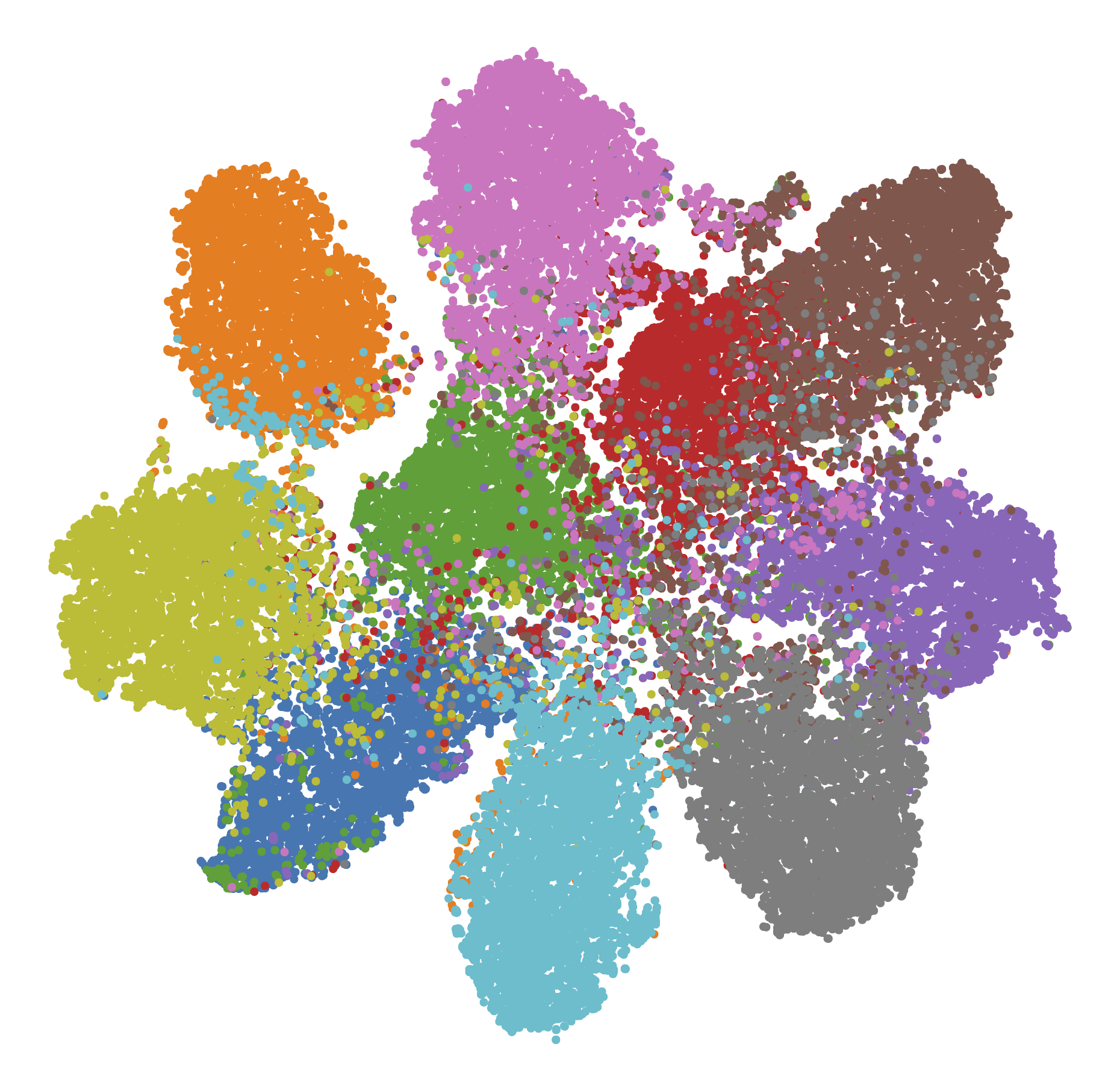} 
\caption{CE}
\end{subfigure}
\begin{subfigure}{0.45\columnwidth}
\centering
\includegraphics[width=1\linewidth]{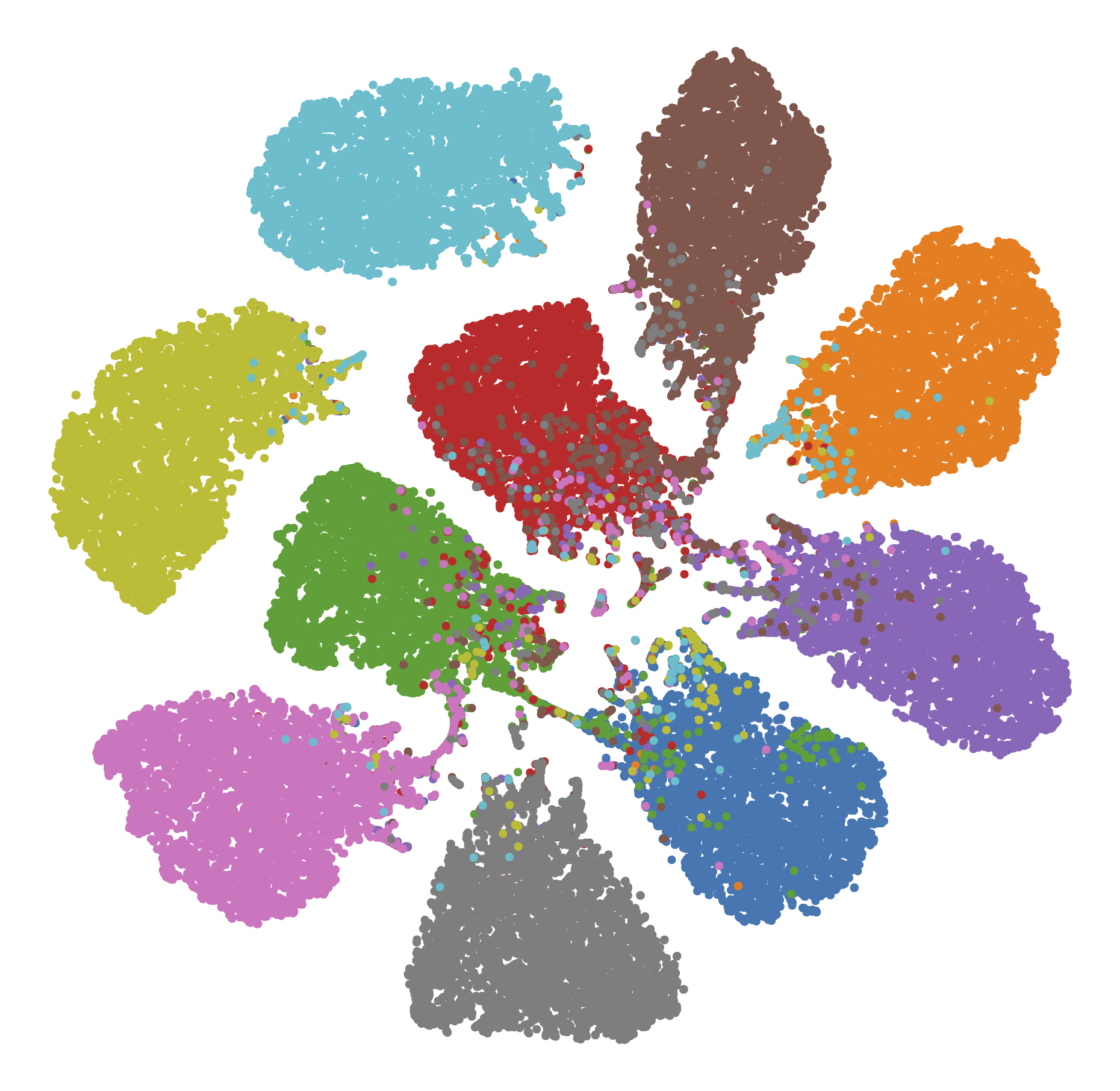} 
\caption{SPR}
\end{subfigure}
\vspace{-0.1in}
\caption{Visualization of learned representations. \label{fig:visualization}
}

\vspace{-0.15in}
\end{figure}

\textbf{Ablation study of modules in SPR}. 
To verify the effectiveness of each module in our framework, we conduct an ablation study on CIFAR10 with 40\% symmetric noise rate. 
Specifically, the ``CE" denotes vanilla cross entropy method; 
the ``CE + SPR" means the cross-entropy loss only on the clean data detected by SPR; 
the ``CE + $\ell_q$" means the~\cref{eq:loss-lq} for all training data;
the ``CE + CutMix'' means using CutMix strategy for all the training data;
other variants are defined similarly based on utilized components,
and the ``Full" denotes our SPR method with all components. 
As shown in~\cref{tab:ablation}, simply using our framework to detect noisy data will lead to better performance compared with the standard CE loss. 
And the full model enjoys the best performance.
We further visualize the learned representation of using SPR compared with standard cross-entropy method in~\cref{fig:visualization}.
SPR will learn a better discriminative representations.

\begin{table}[ht]
\centering
\begin{tabular}{lc}
\toprule
Model & Training Time\tabularnewline
\midrule
\midrule
SPR w/o split algorithm & about 6h\tabularnewline
SPR w/ split algorithm & 54s\tabularnewline
\bottomrule
\end{tabular}
\vspace{-0.1in}
\caption{Training time for one epoch on CIFAR-10.\label{tab:running-time}}

\vspace{-0.1in}
\end{table}

\textbf{Influence of Split algorithm}.
In our framework, we propose a split algorithm to divide the whole training set into small pieces to run SPR in parallel.
In this part, we compare the running time between using the split algorithm and not using it.
Results are shown in~\cref{tab:running-time}.
When we do not use the split algorithm, the training time for each epoch will cost an unacceptable time, making it impossible to train on large datasets.
Hence we propose the split algorithm to reduce the training time.

\begin{figure}

\centering
\includegraphics[width=0.9\linewidth]{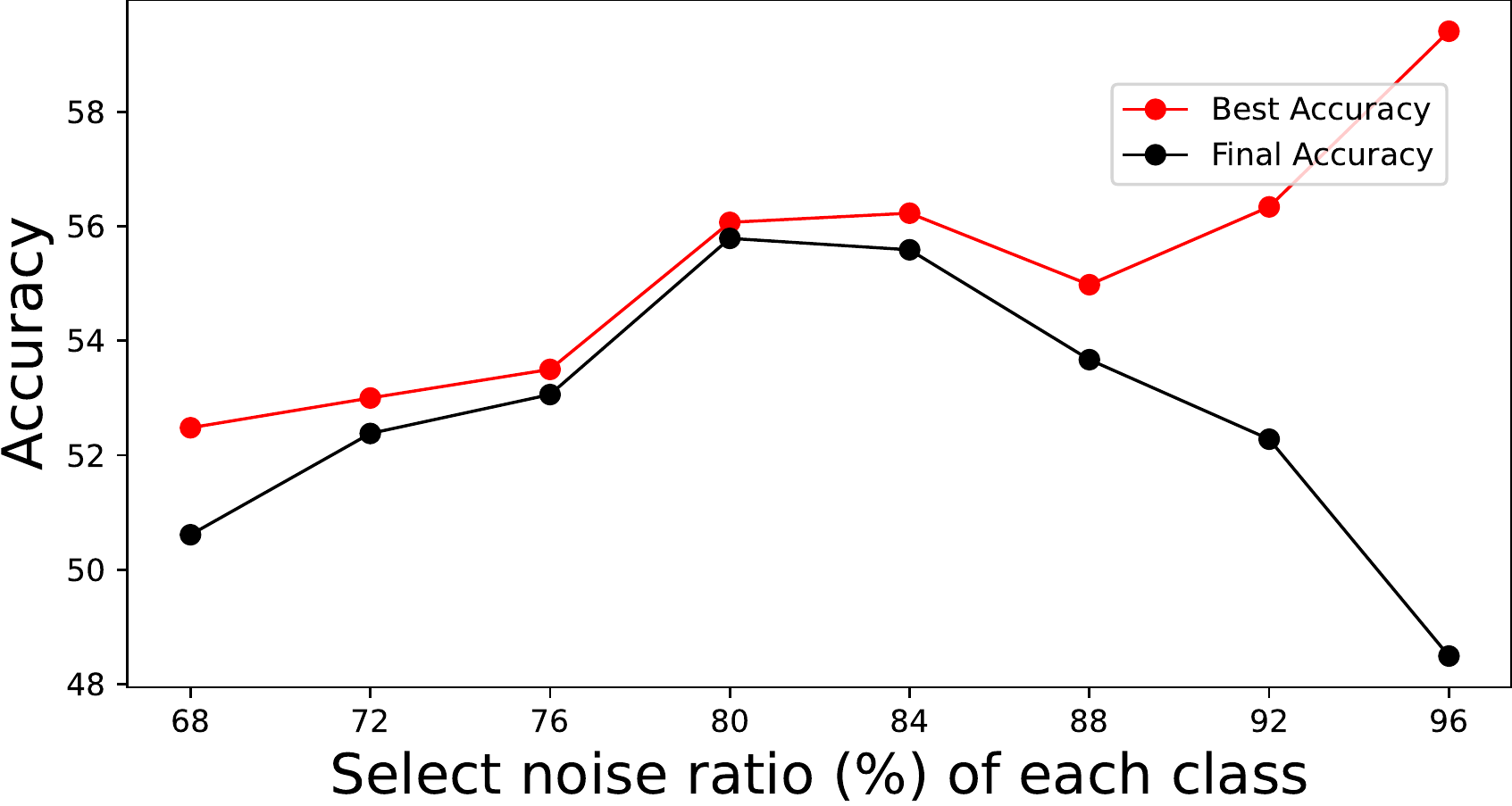}
\vspace{-0.1in}
\caption{Best and final accuracies of SPR running with selecting different ratio of training data. \label{fig:select_num}}

\vspace{-0.1in}
\end{figure}

\textbf{Influence of select ratio}. 
In our experiments, we simply select half of the training data to train the network.
It is desirable to investigate how does the ratio of detected noisy data influence the training process.
We conduct experiments of SPR on CIFAR10 with a symmetric noise rate of 0.8.
To avoid the influence of semi-supervised training pipeline, we only use the supervised training manner in this part.
It can be found that the best selection ratio is near the noise ratio in the training set.
Hence a better selection strategy may be designed based on the estimation of the noise ratio in the training set.
We leave it as a future work since in this paper we mainly propose the sample selection framework.

\begin{figure}
\centering
\includegraphics[width=0.9\linewidth]{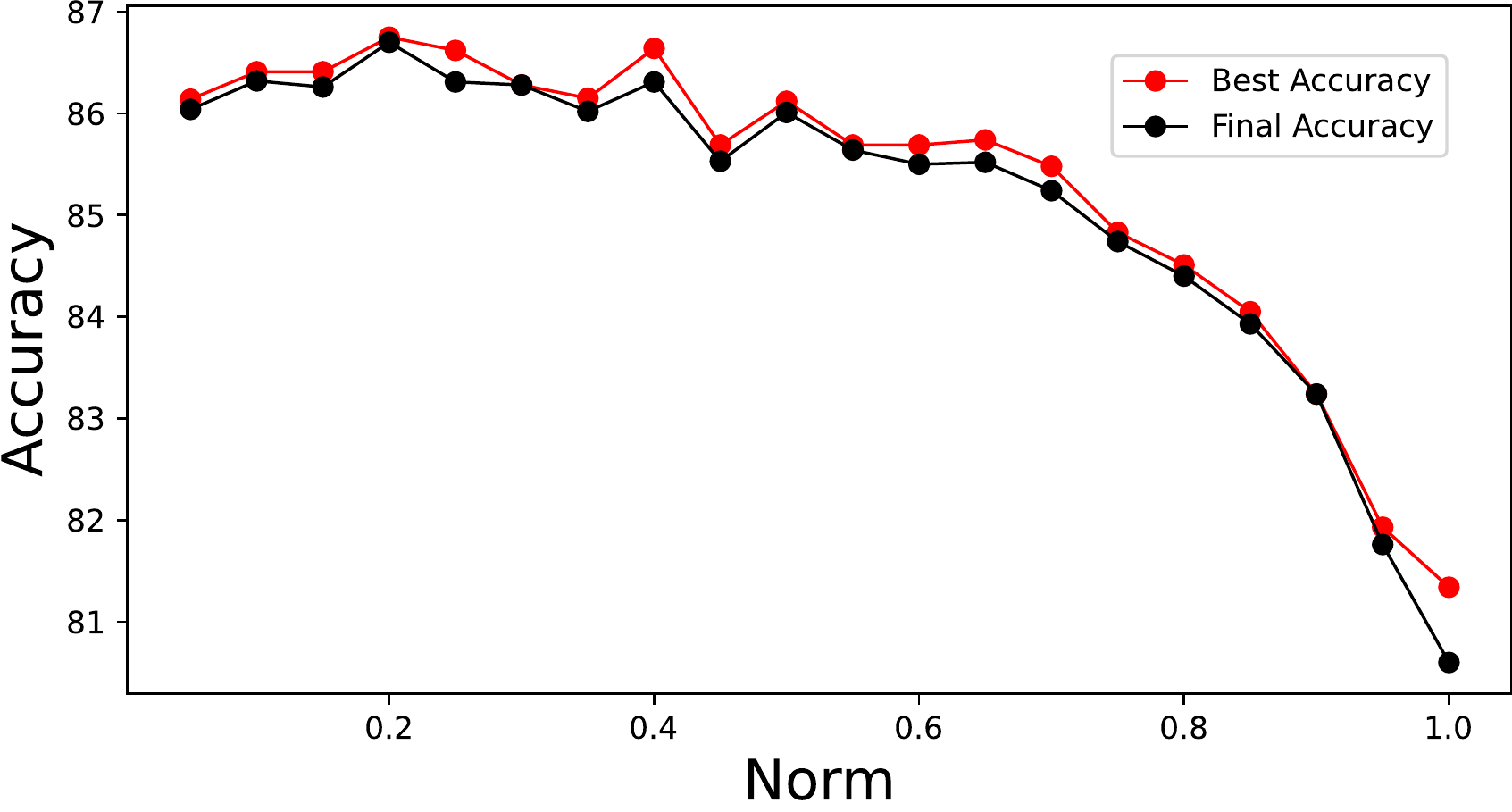}
\vspace{-0.1in}
\caption{Accuracies of SPR with different $\ell_q$. \label{fig:norm}}

\vspace{-0.1in}
\end{figure}

\textbf{Influence of $\ell_q$}. 
In this part, we investigate the influence of $\ell_q$ norm in our framework.
We run with a sequence of $q$ from 0.05 to 1, as illustrated in~\cref{fig:norm}.
In general, a smaller $q$ encourages the linear relation as expected by our framework, while too small $q$ will damage the representation capacity of the network.
Thus, a convex accuracy curve exists when we test with different $\ell_q$, suggesting a choice of $q=0.2$ to be the best.
Hence in our experiments we use $q=0.2$.

\textbf{Limitations of SPR.}
The major limitation of SPR is that it requires the almost necessary irrepresentability condition to identify the noise set.
When this condition is not hold for the problem, SPR will end with a non-vanishing probability to identify at least one clean data as noisy data.
Further, the recovery theory are based on the Gaussian noise assumption, which may not hold for special problems.

\section{Conclusion}
This paper proposes a statistical sample selection framework -- Scalable Penalized Regression (SPR) to identify noisy data with theoretical guarantees.
Specifically, we propose an equivalent leave-one-out t-test approach as a penalized linear model, in which  non-zero mean-shift parameters can be induced as an indicator for  noisy data.
We provide theoretical conditions to guarantee the identifiability of SPR to recover the oracle noisy set.
Experiments on several synthetic  and  real-world datasets show the effectiveness of our method.

\noindent \textbf{Social Impact}. Our SPR will have  positive impact to  social, as it enables to directly identify noisy  data with theoretical grounding to help train the network.

\noindent \textbf{Acknowledgement}. 
This work is supported in part by NSFC under Grant (No. 62076067).

{\small
\bibliographystyle{ieee_fullname}
\bibliography{bib}
}

\end{document}